\documentclass[11pt,a4paper]{article}
\usepackage[hyperref]{eacl2021}
\usepackage{times}
\usepackage{latexsym}

\usepackage{microtype}

\usepackage{subcaption}
\usepackage{multirow,bigdelim}
\usepackage{amsmath}
\usepackage{rotating}
\usepackage{booktabs,xcolor}
\usepackage{xspace}
\usepackage{booktabs}
\usepackage{multirow}
\usepackage{todonotes}
\usepackage{url}
\usepackage{enumitem}
\usepackage[ruled,vlined]{algorithm2e}
\usepackage{enumitem}
\usepackage{amssymb}
\usepackage{amsmath}

\newcommand{\eg}{\textit{e.g.\ }}

\newcommand{\red}[1]{\textcolor{black}{#1}}

\newcommand{\vecb}[1]{\mathbf{#1}}
\newcommand{\T}{\texttt}
\newcommand{\B}{\textbf}
\newcommand{\super}{\texttt{SAC BLEU}\xspace}
\newcommand{\unsup}{\texttt{SAC unsuper}\xspace}

\aclfinalcopy

\title{Exploring Supervised and Unsupervised Rewards in Machine Translation}

\author{Julia Ive$^{1}${\normalfont,} Zixu Wang$^1${\normalfont,}  Marina Fomicheva$^2${\normalfont,} Lucia Specia$^{1,2,3}$ \\[.3em]
Imperial College London$^1$,\, University of Sheffield$^2$,\, ADAPT - Dublin City University$^3$\\
         \texttt{\small j.ive@ic.ac.uk,
         zixu.wang@imperial.ac.uk} \\
         \texttt{\small m.fomicheva@sheffield.ac.uk, l.specia@ic.ac.uk}\\
}

\date{}

\begin{document}
\maketitle
\begin{abstract}
Reinforcement Learning (RL) is a powerful framework to address the discrepancy between loss functions used during training and the final evaluation metrics to be used at test time. When applied to neural Machine Translation (MT), it minimises the mismatch between the cross-entropy loss and non-differentiable evaluation metrics like BLEU.
However, the suitability of these metrics as reward function at training time is questionable: they tend to be sparse and biased towards the specific words used in the reference texts. We propose to address this problem by making models less reliant on such metrics in two ways: (a) with an entropy-regularised RL method that does not only maximise a reward function but also explore the action space to avoid peaky distributions; (b) with a novel RL method that explores a dynamic unsupervised reward function to balance between exploration and exploitation. We base our proposals on the Soft Actor-Critic (SAC) framework, adapting the off-policy maximum entropy model for language generation applications such as MT. We demonstrate that SAC with BLEU reward tends to overfit less to the training data and performs better on out-of-domain data. We also show that our dynamic unsupervised reward can lead to better translation of ambiguous words.
\end{abstract}

\section{Introduction}

Autoregressive sequence-to-sequence (seq2seq) neural architectures have become the {\em de facto} approach in Machine Translation (MT). Such models include Recurrent Neural Networks (RNN)~\citep{NIPS2014_5346, bahdanau2014neural} and Transformer networks~\citep{NIPS2017_7181}, among others. However, these models have as a serious limitation the discrepancy between their training and inference time regimes. They are traditionally trained using the Maximum Likelihood Estimation (MLE), which aims to maximise log-likelihood of a categorical ground truth distribution (samples in the training corpus) using loss functions such as cross-entropy, which are very different from the evaluation metric used at inference time, which generally compares string similarity between the system output and reference outputs. Moreover, during training, the generator receives the ground truth as input and is trained to minimise the loss of a single token at a time without taking the sequential nature of language into account. At inference time, however, the generator will take the previous sampled output as the input at next time step, rather than the ground truth word. MLE training thus causes: (a) the problem of ``exposure bias'' as a result of recursive conditioning on its own errors at test time, since the model has never been exclusively ``exposed'' to its own predictions during
training; (b) a mismatch between the training objective and the test objective, where the latter relies on evaluation using discrete and
non-differentiable measures such as BLEU~\cite{papineni-etal-2002-bleu}.

The current solution for both problems is mainly based on Reinforcement Learning (RL), where a seq2seq
model~\cite{NIPS2014_5346,bahdanau2014neural} is used as the policy which generates actions (tokens) and at each step receives rewards based on a discrete metric taking into account importance of immediate and future rewards. However, RL methods for seq2seq MT models also have their challenges: high-dimensional discrete action space, efficient sampling and exploration, choice of baseline reward, among others~\cite{Choshen2020On}. The typical metrics used as rewards (e.g., BLEU) are often biased and sparse. They are measured against one or a few human references and do not take into account alternative translation options that are not present in the references. 

One way to address this problem is to use entropy-regularised RL frameworks. They incorporate the entropy measure of the policy into the reward to encourage exploration. The expectation is that this leads to learning a policy that acts as stochastically as possible while able to succeed at the task. Specifically, we focus on the Soft Actor-Critic (SAC)~\cite{haarnoja2018soft,haarnoja2018applications} RL framework, which to the best of our knowledge has not yet been explored for MT, as well as other natural language processing (NLP) tasks. The main advantage of this architecture, as compared to other entropy regularised architectures \cite{HaarnojaTAL17,ZiebartMBD08}, is that it is formulated in the off-policy setting that enables reusing previously collected samples for more stability and better exploration. We demonstrate that SAC prevents the model from overfitting, and as a consequence leads to better performance on out-of-domain data.

Another way to address the problem of sparse or biased reward is to design an unsupervised reward. Recently, in Robotics, SAC has been successfully used in unsupervised reward architectures, such as the ``Diversity is All You Need'' (DIAYN) framework~\cite{eysenbach2018diversity}. DIAYN allows the learning of latent-conditioned sub-policies (``skills'') in unsupervised manner, which allows to better explore and model target distributions. Inspired by this work, we propose a formulation of an unsupervised reward for MT. We thoroughly investigate effects of this reward and conclude that it is useful in lexical choice, particularly the rare sense translation for ambiguous words. 

Our {\bf main contributions} are thus twofold: (a) the re-framing of the SAC framework such that it can be applied to MT and other natural language generation tasks (Section~\ref{sec:model}). We demonstrate that SAC results in improved generalisation compared to the MLE training, leading to better translation of out-of-domain data; (b) the proposal of a dynamic unsupervised reward within the SAC framework (Section~\ref{subsec:reward}). We demonstrate its efficacy in translating ambiguous words, particularly the rare senses of such words.
Our datasets and settings are described in Section \ref{sec:exps}, and our experiments in Section \ref{sec:results}.

\section{Related Work} \label{sec:related}

\paragraph{Reinforcement Learning for MT} RL has been successfully applied to MT to bridge the gap between training and testing by optimising the sequence-level objective directly \cite{yu2017seqgan,ranzato2015sequence,bahdanau2016actor}. 
However, thus far mainly the REINFORCE~\cite{williams1992simple} algorithm and its variants have been used \cite{ranzato2015sequence, kreutzer-etal-2018-reliability}. These are simpler algorithms that handle the large natural language action space, but they employ a sequence-level reward which tends to be sparse.

To reduce model variance, Actor-Critic (AC) models consider the reward at each decoding step and use the Critic model to guide future actions \cite{NIPS1999_1786}. This approach has also been explored for MT~\cite{bahdanau2016actor, NIPS2017_6622}. However, more advanced AC models with Q-Learning are rarely applied to language generation problems. This is due to the difficulty of approximating the Q-function for the large action space. The large action space is one of the bottleneck for RL for text generation in general. Pre-training of the agent parameters to be close to the true distribution is thus necessary to make RL work~\citep{Choshen2020On}. Further RL training of the agent makes the overfitting problem even more pronounced resulting in peaky distributions. Such problems are traditionally addressed by entropy regularised RL.

\paragraph{Entropy Regularised RL} The main goal of this type of RL is to learn an efficient policy while keeping the entropy of the agent actions as high as possible. The paradigm promotes exploration of actions, suppresses peaky distributions and improves robustness. In this work, we explore the effectiveness of the maximum entropy SAC framework~\cite{haarnoja2018soft}. 

The work closest to ours is of~\citet{dai2018credit} where the Entropy-Regularised AC (ERAC) model leads to better MT performance. The major difference between ERAC and SAC is that the former is an on-policy model and the latter is an off-policy model. On-policy approaches use consecutive samples collected in real-time that are correlated to each other. In the off-policy setting, our SAC algorithm uses samples from the memory that are taken uniformly with reduced correlation. This key characteristic of SAC ensures better model generalisation and stability~\cite{mnih-dqn-2015}. There are also differences in the architectures of SAC and ERAC, i.a., using 4 Q-value networks instead of two. These differences will be covered in detail in Section~\ref{sec:model}.

\paragraph{Unsupervised reward RL} 
Significant work has been done in Robotics to improve the learning capability of robots. These approaches do not rely on a single objective but rather promote intrinsic motivation and exploration. Such an approach to learn diverse skills (latent-conditioned sub-policies, in practice, skills like walking or jumping) in unsupervised manner was recently proposed by~\citet{eysenbach2018diversity}. The approach relies on the SAC model and inspired our approach to designing our unsupervised reward for MT. We are not aware of other attempts to design dynamic unsupervised RL rewards (learnt together with the network) in seq2seq in general, or MT in particular. Recent work on unsupervised rewards in NLP~\cite{gao-etal-2020-supert} explores mainly static rewards computed against synthetic references.

\section{Methodology}\label{sec:model}

In this section we start by describing the underlying MT architecture and its variant using RL, to then introduce our SAC formulation and the reward functions used.

\subsection{Neural Machine Translation (NMT)}
 
A typical Neural Machine Translation (NMT) system is a seq2seq architecture~\cite{NIPS2014_5346, bahdanau2014neural}, where each source sentence $\vecb{x} = (x_{1}, x_{2}, \cdots, x_{n})$ is encoded by the encoder into a series of hidden states. At each decoding step $t$, a target word $y_{t}$ is generated according to $p(y_{t}|y_{<t}, x)$ conditioned on the input sequence $x$ and decoded sequence $\vecb{y}_{<t} = (y_{1}, \cdots, y_{t-1})$ up to the $t$-th time step. 
Given the corpus of pairs of source and target sentences $\{x_{i}, y_{i}\}_{i=1}^{N}$, the training objective function - maximum likelihood estimation (MLE) is defined as:
\begin{equation}
    \mathcal{L}_{\textrm{MLE}} = - \sum_{i=1}^{N}\sum_{t=1}^{T} p(y_{t}^{i}|y_{1}^{i}, ..., y_{t-1}^{i}, x^{i})
\end{equation}

\subsection{Reinforcement Learning for NMT}

Within the RL framework, the task of NMT can be formulated as a sequential decision making process, where the \textbf{state} is defined by the previously generated words ($\vecb{y}_{<t}$) and the \textbf{action} is the next word to be generated. Given the state $s_t$, the agent picks an action $a_t$ (for seq2seq it is the same as $y_t$), according to a (typically stochastic) \textbf{policy} $\pi_{\theta}$ and observes a reward $r_t$ for that action. The reward can be calculated based on any evaluation metric, \eg BLEU.

The objective of the RL training is to maximise the expected reward:
\begin{equation}
\mathcal{L}_{RL} = \mathop{\mathbb{E}_{a_1,\cdots,a_T \sim {\pi_{\theta}(a_1,\cdots,a_T)}}}[r(a_1,\cdots,a_T)]
\label{eq:rl}
\end{equation}

Under the policy $\pi$, we can also define the values of the state-action pair $Q(s_t, y_t)$ and the state $V(s_t)$ as follows:
\begin{equation}
\begin{array}{l}
Q_{\pi}(s_t,a_t)=\mathop{\mathbb{E}} [r_{t}|s=s_{t},a=a_t]\\
V_{\pi}(s_t) = \mathop{\mathbb{E}_{a \sim \pi(s)}}[Q_{\pi}(s_t,a=a_t)]
\end{array}
\label{eq:qv}
\end{equation}

Intuitively, the value function $V$ measures how good the model could be when it is in a specific state $s_t$. The $Q$ function measures the value of choosing a specific action when we are in such state.

Given the above definitions, we can define a function called {\em advantage} -- denoted by $A_{\pi}$ -- relating the value function $V$ and $Q$ function as follows:
\begin{equation}
A_{\pi}(s_{t},a_t) = Q_{\pi}(s_{t},a_t)-V_{\pi}(s_{t})
\label{eq:adv}
\end{equation}

Therefore, the focus is on maximising one of the following objectives:

\begin{equation}
\text{max}_{a}\ A_\pi(s_t, a_t) \rightarrow \text{max}_{a}\ Q_\pi(s_t, a_t)
\label{eq:qargmax}
\end{equation}

Different RL algorithms have different ways to
search for the optimal policy. Algorithms such as REINFORCE, as well as its variant MIXER~\cite{ranzato2015sequence}, popular in language tasks, search for the optimal policy via~Eq.~\ref{eq:rl} using the Policy Gradient. Actor-Critic (AC) models typically improve the performance of Policy Gradient models by solving Eq.~\ref{eq:qargmax} (left part)~\cite{bahdanau2016actor}. $Q$-learning models that aim at maximising the $Q$ function (Eq~\ref{eq:qargmax}, right part) to improve over both the Policy Gradient and AC models~\cite{dai2018credit}. 
 
\subsection{Soft Actor-Critic (SAC)} \label{ssec:sac}

The SAC algorithm \cite{haarnoja2018soft} adds to the Eq.~\ref{eq:rl} an entropy term:
\begin{equation}
    \mathcal{L}(\pi) = \sum_{t=1}^{T} \underset{a_t\sim \pi_{(\cdot | s_t)}}{\mathbb{E}} [r(s_t, a_t) + \alpha \mathcal{H} (\pi (\cdot | s_t))]
\label{eq:ziebart-entropy}
\end{equation}
where $\alpha$ controls the stochasticity of the optimal policy, a trade-off between the relative importance of the entropy term $\mathcal{H}$ and the reward $r(s_t, a_t)$  that the agent receives by taking action $a_t$ when the state of the environment is $s_t$. Its aim is to maximise the entropy of actions at the same time as maximising the rewards. 

As mentioned earlier, SAC is an off-policy $Q$-learning AC algorithm. As other AC algorithms it consists of two parts: the actor (the policy function) and the critic -- action-value function (Q), parameterised by $\phi$ and $\theta$, respectively.
 
During off-policy learning, the history of states, actions and respective rewards are stored in a memory ($D$), {\em a.k.a.} the replay buffer. 

\begin{itemize}[leftmargin=0cm,itemindent=.5cm,labelwidth=\itemindent,labelsep=0cm,align=left]
\item \textbf{Critic Training}

The Q-function estimates the value of an action at a given
state based on its future rewards. The soft-Q value is computed recursively by applying a modified Bellman backup operator: 
\begin{equation}
    Q(s_t, a_t) = r(s_t, a_t) + \gamma \underset{s_{t+1} \sim D}{\mathbb{E}} [V(s_{t+1})] 
\label{eq:soft-q-value}
\end{equation}
\noindent where 
\begin{equation}
V(s_t) = \underset{a_t \sim \pi}{\mathbb{E}} [Q(s_t, a_t)-\alpha \log \pi (a_t | s_t)]
\label{eq:soft-state-value-func}
\end{equation}
is the expected future reward of a state and $\log(\pi(a_{t}|s_{t}))$ is the entropy of the policy.  

The parameters of the Q-function are updated towards minimising the mean squared error between the estimated Q-values and the assumed ground-truth Q-value. The assumed ground-truth Q-values are estimated based on the
current reward ($r(s_t, a_t)$) and the discounted future reward of the next state ($\gamma V_{\bar\theta} (s_{t+1})$). This mean squared error objective
function of the Q network is as follows:

\begin{equation}
\begin{split}
    \mathcal{L}(\theta) = \underset{s_t, a_t, r_t, s_{t+1} \sim D, a_{t+1}\sim \pi_{\phi}}{\mathbb{E}} \Big[\big(Q_\theta (a_t, s_t) - \\ [r(s_t,a_t)
    + \gamma \underset{s_{t+1} \sim D}{\mathbb{E}} [V_{\bar{\theta}}(s_{t+1})]]\big)^2 \Big]
\end{split}
\label{eq:mse-loss}
\end{equation}

Note that the parameters of the networks are denoted as $\theta$ and $\bar\theta$ respectively. This is the best practice where the critic is modeled
with two neural networks with the exact same architecture but independent parameters~\cite{mnih-dqn-2015}. 

The parameters of the target critic network ($Q_{\bar\theta}$) are iteratively updated with the exponential moving average of the parameters of the main critic network ($Q_\theta$). This constrains the parameters of the target network to update at a slower pace toward the parameters of the main critic, which has been shown to stabilise the training process \cite{lillicrap16}.

Another advantage of SAC is the double Q-learning \cite{hasselt10}. In this approach,
two Q networks for both of the main and the target critic
functions are maintained. When estimating the current Q values or the discounted future rewards, the minimum of the outputs of the two Q networks is used. Thus the estimated Q values do not grow too large,  which improves the policy training \cite{haarnoja2018soft}.

\item{\textbf{Actor Training}}

SAC updates the policy to minimise the KL-divergence to make the distribution of $\pi_{\phi}(s_{t})$ policy function look more like the distribution of the Q function:
\begin{equation}
    \mathcal{L}_{\pi}(\phi) = \underset{s_t \sim D}{\mathbb{E}}[\pi_{t}(s_{t})^{T}[\alpha\log(\pi_{\phi}(s_{t})) - Q_{\theta}(s_{t})]]
\label{eq:erac_actor}
\end{equation}
where softmax is used in the final layer of the policy to output a probability distribution over the actions.
\end{itemize}

We note that some versions of the SAC algorithm allow to automatically tune the $\alpha$ parameter so that while maximising the expected return, the policy should satisfy the minimum entropy criteria. In our experiments we however used a fixed $\alpha$. Updating $\alpha$ during training resulted in too short sentences in the output.

Finally, we note that Eq.~\ref{eq:erac_actor} does \textit{not} simply add an entropy term to the standard Policy Gradient.
The critic $Q_\theta$ trained by Eq.~\ref{eq:mse-loss} additionally captures the \textit{entropy from future steps}.

For more details on SAC for the discrete setting (like MT)  we refer to \citet{christodoulou2019soft}. For more formal details on the architecture, see~\citet{haarnoja2018soft,haarnoja2018applications}.

\subsection{Reward functions}\label{subsec:reward}

Below we define the reward functions we use in our SAC architecture. 

\paragraph{Supervised BLEU reward: - \super}
In the supervised setup, we employ the sequence-level BLEU score~\cite{papineni-etal-2002-bleu} with add-1 smoothing~\cite{chen-cherry-2014-systematic}. As an additional length constraint at each time step, we deduct from the respective score the length penalty: $lp = |l_{\vecb{y}} -l_{\hat{\vecb{y}}}|$, where $\vecb{y}$ is the reference translation. This penalty prevents longer translations that are not penalised by the brevity penalty of BLEU. BLEU has been chosen in our study to ensure better comparability with the related work in RL MT traditionally using the BLEU reward~\cite{bahdanau2016actor,dai2018credit}. 

\paragraph{Unsupervised reward - \unsup}
As discussed above, using automatic metrics as reward function can lead to a number of issues, \eg reward sparsity, overfitting towards single reference. Moreover, designing a good reward can be challenging.

Inspired by recent work on the SAC algorithm in unsupervised RL~\cite{eysenbach2018diversity}, we have designed an unsupervised reward that \textit{balances the quality and diversity in the model search space}.

The pseudo-reward function we use is as follows:
\begin{equation}
   r_z(\vecb{x},a) = \log q_\delta (z \vert \vecb{x}, a) - \log p(z)
\end{equation}
where $p(z)$ is a categorical uniform distribution for a latent variable $z$. 

$q_\delta(z \vert \vecb{x}, a)$ is provided by a discriminator parametrised by a neural network. $z$ is randomly assigned to a word sampled at each step from the actor distribution. The discriminator is a Bag-of-Words model that takes as input the encoded source sequence and the word itself to predict its $z$. 

More intuitively, every time a word appears in the translation hypothesis for a source sentence (within the Bag-of-Words formulation) it is randomly assigned a certain value of $z$. The more times this word appears in the sampled hypotheses (for a given source) the closer will be $\log q_\delta (z \vert \vecb{x}, a)$ to the uniform prior $p(z)$, hence reward $r_z(\vecb{x}, a)$ will be close to 0. Thus, frequent translations will be suppressed and search for less frequent translations will be encouraged in order to receive a reward larger than 0.
 
Such a reward is less sparse than the traditional ones and is also dynamic which prevents memorising and overfitting. 

\section{Experimental Setup}\label{sec:exps}
\subsection{Data}

We perform experiments on the \B{Multi30K} dataset~\cite{elliott-etal-2016-multi30k}\footnote{https://github.com/multi30k/dataset} of image description translations and focus on the English-German (EN-DE) and English-French (EN-FR)~\cite{elliott-etal-2017-findings} language directions. Following best practises, we use sub-word segmentation (BPE~\cite{sennrich-etal-2016-neural}) only on the target side of the corpus. The dataset contains 29,000 instances for training, 1,014 for development, and 1,000 for testing. We use \B{flickr2016} (\B{2016}), \B{flickr2017} (\B{2017}) and \B{coco2017} (\B{COCO}) test sets for model evaluation. 
 
\B{2016} is the most \textbf{in-domain} test set since it was taken from the same superset of descriptions as the training set, whereas \B{2017} and \B{COCO} are from different image description corpora and are thus considered \textbf{out-of-domain}. 

For more fine-grained assessment of our models with unsupervised reward, we use the \B{MLT} test set~\cite{LALA18.629, lala-etal-2019-grounded}, an annotated subset of the \B{Multi30K} corpus where each instance is a 3-tuple consisting of an \B{ambiguous} source word, its textual context (a source sentence), and its correct translation. The test set contains 1,298 sentences for English-French and 1,708 for English-German. It was designed to benchmark models in their ability to select the right lexical choice for words with multiple translations, especially when some of these translations are rarer.

Additionally, to allow for comparison with previous work, we evaluate on the \textbf{IWSLT 2014} German-to-English dataset~\cite{mauro2012wit3} from TED talks, which has been used as testbed in most work on RL for MT. The training set contains  $153K$ sentence pairs. We followed the pre-processing procedure described in~\cite{dai2018credit}. 

When compared to the \B{IWSLT 2014} dataset, all the three \B{Multi30K} test sets are more out-of-domain. This was found by the analysis of perplexities of language models trained with respective training data for each dataset (see Appendix~\ref{appssec:domain}).

\begin{table*}[!ht]
\begin{center}
\scalebox{0.84}{
\begin{tabular}{lc@{\hspace{1cm}}lll@{\hspace{0.5cm}}lll@{\hspace{0.5cm}}lll@{\hspace{0.5cm}}}
\toprule
& & \multicolumn{3}{c}{\B{2016}} & \multicolumn{3}{c}{\B{2017}} & \multicolumn{3}{c}{\B{COCO}} \\ 
\cline{3-11}
& model & BLEU & METEOR & TER & BLEU & METEOR & TER & BLEU & METEOR & TER \\ \midrule
\multirow{4}{*}{\rotatebox{90}{EN-FR}}
& MLE & 57.5 & 71.7 & 27.5 & 50.9 & 66.8 & 33.0 & 42.8 & 61.5 & 37.3 \\
& ERAC (ours) & \bf 59.4* & \bf 73.3* & \bf 26.7* & 51.2 & 66.8 & 32.5 & 42.5* & 60.6* & 37.6* \\
& \super & 57.9 & 72.0 & 27.8 & \bf 51.7* & \bf 67.5* & \bf 32.1* & \bf 44.4 & \bf 62.9 & \bf 36.4 \\
& \unsup & 56.9 & 71.4 & 28.2 & 51.1 & 67.1 & 32.5 & 43.6 & 62.6* & 36.6\\
\midrule
\multirow{4}{*}{\rotatebox{90}{EN-DE}}
& MLE & 38.5 & \bf 57.2 & 42.2 & \bf 31.9 & \bf 51.3 & 49.5 & 27.2 & 46.7 & 55.2 \\
& ERAC (ours) & \bf 38.9* & 56.1* & \bf 41.9* & 31.4* & 49.6* & 49.7* & 25.0 & 44.0 & 56.0* \\
& \super & 38.1 & 56.8* & 42.5 & \bf 31.9 & 51.2 & \bf 49.1 & \bf 27.7 & \bf 47.0 & \bf 54.5\\
& \unsup & 38.0* & 56.9 & 43.0* & 31.6 & 50.8 & 49.7* & 26.6 & 46.5 & 55.1\\
\bottomrule
\end{tabular}
}
\end{center}
\caption{Performance of \super on the \B{Multi30K} test sets (EN-FR, EN-DE) trained on the \B{Multi30K} train set. * marks statistically significant changes (p-value $\leq 0.05$) as compared to MLE. Bold highlights best results. ERAC (ours) indicates results obtained by us using the code openly provided by~\citet{dai2018credit}.}
\label{tab:res_multi30k}
\end{table*}

\subsection{Training}

 We modify the original SAC architecture to adapt it to MT following best practices~\cite{bahdanau2016actor} in the area. The functions $\pi_\phi$ and $Q_\theta$ are parameterised with neural
networks: $\pi_\phi$ is an RNN seq2seq model with a 2-layer GRU~\cite{cho-etal-2014-learning} encoder and a 2-layer Conditional GRU decoder~\cite{sennrich-etal-2017-nematus} with attention~\cite{bahdanau2014neural}. For \super, $Q_\theta$ duplicates the structure of the former, but encodes the reference instead of the source sentence to mimic inputs to the actual BLEU function.

We first pretrain the actor and then pretrain the critic, before the actor-critic training. The pretraining of actors is done until convergence according to the early stopping criteria of 10 epochs wrt. to the MLE loss. We have also found that our critics require much less pretraining (3-5 epochs as compared to 10-20 epochs in general for AC architectures with the MSE loss).
Also, to prevent divergence during the actor-critic training, we continue performing MLE training using a smaller weight $\lambda_\text{mle}$. 
We set $\alpha$ to 0.01. Following~\citet{haarnoja2018soft}, we rescale the reward to the value inverse to $\alpha$. Note that we did not find it useful to add to SAC the smoothing objective minimising variance of Q-values~\cite{bahdanau2016actor,dai2018credit}. We presume that the double Q-learning significantly contributes to the stability of the network and additional smoothing is not required. 

For \unsup, we parameterise $q_\delta$ by a 2-layer feed-forward neural network, which takes the source as encoded by the actor and $a_t$ and outputs $q_\delta (z \vert \vecb{x}, a)$. We set $z$ to take one of 4 values.\footnote{This hyperparameter is tuned on the validation set. It typically varies from 2 to several hundreds in the related work~\cite{haarnoja2018applications}.} For this unsupervised setting, we do not train a Q-function. We instead operate in the oracle mode and following~\cite{keneshloo2018deep} define true Q-value estimates and use it to update our actor. Details on training are given in Appendix~\ref{appsec:training}. We use \T{pysimt}~\cite{caglayan-etal-2020-simultaneous} with PyTorch~\cite{paszke2017automatic} v1.4 for our experiments.\footnote{https://github.com/ImperialNLP/pysimt}

\begin{table*}[!ht]
\begin{center}
\scalebox{0.85}{
\begin{tabular}{lc@{\hspace{1cm}}lll@{\hspace{0.5cm}}lll@{\hspace{1.2cm}}lll@{\hspace{0.5cm}}}
\toprule
& & \multicolumn{3}{c}{\B{2016}} & \multicolumn{3}{c}{\B{2017}} & \multicolumn{3}{c}{\B{COCO}} \\ 
\cline{3-11}
& Model & BLEU & METEOR & TER & BLEU & METEOR & TER & BLEU & METEOR & TER \\ \midrule
\multirow{2}{*}{\rotatebox{90}{\small{UNK}}}
& MLE & 25.1 & \bf 29.1 & \bf 49.9 & 23.1 & \bf 27.7 & 54.5 & 18.9 & \bf 25.8 & 59.2 \\
& \super & \bf 25.2 & 28.9 & 50.1 & \bf 23.2 & 27.5 & 54.5 & \bf 19.4 & 25.5* & \bf 58.3*  \\ \midrule
\multirow{2}{*}{\rotatebox{90}{\small{noUNK}}}
& MLE & 34.4 & 37.8 & 40.4 & 31.6 & \bf 37.7 & 46.4 & 25.9 & 34.2 & 50.6 \\
& \super & \bf 34.9 & \bf 38.0 & \bf 40.0 & \bf 32.1 & \bf  37.6 & \bf 45.9 & \bf 28.3* & \bf 34.5 & \bf 48.6 \\ 
\bottomrule
\end{tabular}
}
\end{center}
\caption{Performance of \super on \B{Multi30K} (German-English) trained on the \B{IWSLT 2014} train set. UNK indicates standard output containing the UNK symbol; noUNK -- outputs with sentences containing UNK not taken into account. * marks statistically significant changes (p-value $\leq 0.05$) as compared to MLE. Bold highlights best results.}
\label{tab:res_multi30k-iwslt}
\end{table*}

\subsection{Evaluation}

We use the standard set of MT evaluation metrics: BLEU~\cite{papineni-etal-2002-bleu}, METEOR~\cite{denkowski:lavie:meteor-wmt:2014} and TER~\cite{snover-AMTA-2006}.
We perform significance testing via bootstrap resampling using the \texttt{Multeval} tool \cite{clark-etal-2011-better}.

For the lexical translation task, we measure the \textbf{Lexical Translation Accuracy (LTA)} score \cite{lala-etal-2019-grounded}. The score provides an average estimation of how accurately the words have been translated. For each ambiguous word, a score of +1 is awarded if the correct translation of the word is found in the output translation; a score of 0 is assigned if a known incorrect translation is found, or none of the candidate words are found in the translation.
We also propose a metric that not only rewards correctly translated ambiguous words, but also penalises words translated with the wrong sense: the \textbf{Ambiguous Lexical Index (ALI)}. ALI assigns -1 for wrong translations in the given context, whereas LTA simply does not reward them.

\section{Results} \label{sec:results}

\subsection{Comparison to state-of-the-art} \label{ssec:results-sota}

We first compare our SAC models against the MLE model (baseline) and ERAC\footnote{For ERAC, we present results that we reproduced ourselves using the code publicly provided by the authors. We had to perform several modifications to this code to make it conform  recent deep learning framework software updates. The performance of this model is on pair with this reported by the authors.} (state-of-the-art -- SOTA) both trained and tested on the \B{Multi30K} data (Table~\ref{tab:res_multi30k}). 
Compared to SAC, ERAC differs in that it uses the on-policy setting (i.e., using samples collected in real time). Our SAC algorithm is an off-policy algorithm and uses samples from the memory to promote generalisation.

We clearly observe the tendency of ERAC models to perform better on the more in-domain \B{2016} data (+1.9 BLEU, +1.6 METEOR, -0.8 TER against MLE for EN-FR) and the tendency of \super models to outperform other models on more out-of-domain \B{2017} and \B{COCO} sets (+2.7 BLEU and +3.0 METEOR, -1.5 TER against ERAC on \B{COCO} for EN-DE). 

\unsup results are however worse than the baseline and SOTA. We focus thus on the investigation of \super and come back to \unsup in Section~\ref{ssec:results-amb}.

To further confirm our hypothesis that SAC reduces overfitting and performs better on the out-of-domain data, we train our models on the \B{IWSLT 2014} train set and test on the out-of-domain \B{Multi30K} test sets (in the reverse direction, German into English, Table~\ref{tab:res_multi30k-iwslt}).

We observe similar performance for complete set of outputs (including sentences with UNK tokens) for MLE and \super. If the lines with UNK words are not taken into account,\footnote{The original corpus pre-processing pipeline that we followed to increase comparability does not include subword segmentation. \red{We take the intersection of hypotheses sentences across \B{Multi30K} test setups that contain no generated UNK token wrt. the \B{IWSLT 2014} vocabulary. Reference files may still contain the UNK token, we focus on the generated text here.}}  we observe an improvement for the \B{2016} and \B{2017} test sets (+0.5 BLEU, +0.1 METEOR, -0.5 TER on average), and a much bigger improvement for the more out-of-domain \B{COCO} set (+2.5 BLEU, +0.3 METEOR, -2 TER on average). This confirms our hypothesis that SAC helps to reduce overfitting.

Finally, we compare SAC to the SOTA AC-base RL architectures, namely ERAC and AC, on the \B{IWSLT 2014} set that is commonly used for this task.
Compared to SAC, AC differs in that it does not use entropy regularisation. We also provide the performance for the popular MIXER algorithm. Results are shown in Table~\ref{tab:iwslt}. 

In terms of the general performance, our SAC performs on pair with the MLE model. \super even slightly lowers this score (-0.2 BLEU, -0.2 METEOR). We note that \super results contain an increased count of UNK words as compared to MLE (+2.8\%) This increased generation of UNK words due to the entropy regularisation is partially responsible for this similar performance. Another cause is that SAC does not overfit to the BLEU distribution of the target data.\footnote{\red{We mean that the model would have a tendency to select certain words to simply boost BLEU rather than picking words to reflect the correct meaning.}} 

\begin{table}[!ht]
	\centering
	\scalebox{0.73}{
	\begin{tabular}{l@{\hspace{0.2cm}}lll}
		\toprule
		Model & BLEU & METEOR & TER \\
		\midrule
		MLE (ours) & 29.8 & 31.2 & 48.9 \\
		\cline{1-4}
		MIXER~\cite{ranzato2015sequence} & 20.73 & - & - \\
		AC~\cite{bahdanau2016actor} & 28.53 & - & - \\
		ERAC (w/feed)~\cite{dai2018credit} & 29.36 & - & - \\
		ERAC (w/o feed)~\cite{dai2018credit} & 28.42 & - & - \\
		ERAC (w/o feed, ours) & 29.0* & 30.6* & 51.5* \\
		\midrule
		\super & \bf 29.6* & \bf 31.0* & \bf 48.8*  \\
		\bottomrule
	\end{tabular}}
	\caption{Performance of MLE and different RL algorithms on the \B{IWSLT 2014} test set trained on the \B{IWSLT 2014} train set. * marks statistically significant changes (p-value $\leq 0.05$) as compared to MLE. Bold highlights best RL results. MIXER, AC and ERAC scores were taken from original papers. ERAC (ours) indicates our results using the code provided in~\cite{dai2018credit}.}
	\label{tab:iwslt}
\end{table}

\begin{table*}[!ht]
\begin{center}
\scalebox{0.77}{
\begin{tabular}{lc@{\hspace{1cm}}lc@{\hspace{0.5cm}}lc@{\hspace{0.5cm}}lc@{\hspace{0.5cm}}}
\toprule
& & \multicolumn{6}{c}{\textbf{All Cases}} \\
& & \multicolumn{2}{c}{\B{2016}} & \multicolumn{2}{l}{\B{2017}} & \multicolumn{2}{l}{\B{COCO}} \\ 
\cline{3-8}
& Model & LTA & ALI & LTA & ALI & LTA & ALI \\ \midrule
\multirow{3}{*}{\rotatebox{90}{EN-FR}}
& MLE & 81.60 & 63.19 & 79.65 & 59.31 & 74.60 & 49.21  \\
& \super & 81.94 & 63.89 & 79.76 & 59.53 & \bf 77.32 & \bf 54.65\\
& \unsup & \bf 82.75 & \bf 65.51 & \bf 80.62 & \bf 61.25 & 75.28 & 50.57 \\\midrule
\multirow{3}{*}{\rotatebox{90}{EN-DE}}
& MLE & 65.34 & 30.68 & 70.91 & 41.82 & 67.45 & 34.91 \\
& \super & 64.74 & 29.48 & 71.93 & 43.86 & \bf 67.72 & \bf 35.43 \\
& \unsup & \bf 65.54 & \bf 31.08 & \bf 73.41 & \bf 46.82 & 66.40 & 32.81 \\
\bottomrule
\end{tabular}

\begin{tabular}{lc@{\hspace{0.5cm}}lc@{\hspace{0.5cm}}lc@{\hspace{0.5cm}}}
\toprule
\multicolumn{6}{c}{\textbf{Rare Cases}} \\
\multicolumn{2}{c}{\B{2016}} & \multicolumn{2}{l}{\B{2017}} & \multicolumn{2}{l}{\B{COCO}} \\ 
\cline{1-6}
LTA & ALI & LTA & ALI & LTA & ALI \\ \midrule
52.81 & 24.49 & \bf 47.80 & \bf 16.48 & 47.16 & \bf 18.49  \\
53.37 & 25.39 & 45.91 & 13.46 & \bf 49.05 & 15.47  \\
\bf 54.49 & \bf 27.19 & \bf 47.80 & \bf 16.48 & 47.16 & 15.47  \\\midrule
50.95 & 11.72 & 60.00 & 28.00 & 56.56 & 21.82 \\
50.14 & 10.24 & 60.58 & 29.04 & \bf 58.58 & \bf 25.45 \\
\bf 51.50 & \bf 12.70  & \bf 63.77 & \bf 34.78 & 52.52 & 14.55 \\
\bottomrule
\end{tabular}
}
\end{center}
\caption{Performance of \super on the \B{MLT} test sets (EN-FR, EN-DE). We report Ambiguous Words Accuracy: LTA and ALI. \textbf{Rare Cases} indicates the cases where the correct translation is {\em not} the most frequent translation in the training set.}
\label{tab:lex_translation}
\end{table*}

\subsection{Translation of ambiguous words}\label{ssec:results-amb}

To further investigate the effect of the unsupervised reward, we have evaluated \unsup on the \B{MLT} dataset. Results are shown in Table~\ref{tab:lex_translation}. We calculate the scores on two conditions: \textbf{All Cases} takes into account all possible lexical translations; while for \textbf{Rare Cases}, only the instances where the gold-standard translation is not the most frequent translation for that particular ambiguous word. We observe that both \super and \unsup outperform the MLE baseline across metrics in all setups except for the \B{COCO} EN-FR translation in Rare Cases, where MLE performs better. For \super, this observation is also shown by general evaluation metrics BLEU, METEOR and TER on all \B{MLT} test sets (see Table~\ref{table:amb-bleu} in Appendix).

Moreover, \unsup is particularly successful when evaluated on \B{2016} and \B{2017} and outperforms both MLE and \super across setups. This demonstrates the potential of the unsupervised reward function for the cases when we have to choose between possible translations for an ambiguous word (i.e., better exploration of the search space). BLEU reward, on the other hand, is more reliable when we have to adjust distributions to produce one single possible translation. Manual inspection of these \unsup improvements confirmed their increased accuracy (see Table~\ref{tab:amb-samples}). For example, the ambiguous French source word `hill' (`colline') is translated as `pente'(`slope') by both MLE and \super, while only \unsup produces the correct sentence: `adolescent saute la \textit{colline} `hill' avec son vélo'.
 
 \begin{table*}[t!]
\begin{center}
\scalebox{1.0}{
\small{
\begin{tabular}{c||l|p{11.5cm}}
\hline
EN-FR & source word & hill \\
   & gold target word & colline\\
   & source sentence & the teen jumps the \textbf{hill} with his bicycle .\\
   & reference sentence & ado saute sur la \textbf{colline} `hill' avec son vélo .\\
\cline{2-3}
   & MLE & adolescent saute sur la \textbf{pente} `slope' avec son vélo .\\
   & \super & adolescent saute la \textbf{pente} `slope' avec son vélo .\\
   & \unsup & adolescent saute la \textbf{colline} `hill' avec son vélo .\\
\hline
EN-DE & source word & outfit\\
   & gold target word & outfit\\
   & source sentence & a rhythmic gymnast in a blue and pink \textbf{outfit} performs a ribbon routine .\\
   & reference sentence & eine rhythmische sportgymnastin in einem blauen und pinken \textbf{outfit} vollführt eine bewegung mit dem band .\\
\cline{2-3}
   & MLE & ein begeisterter turner in blau-rosa \textbf{kleidung} `dress' führt eine band auf .\\
   & \super & ein begeisterter turner in blau-rosa \textbf{kleidung} `dress' führt eine band auf .\\
   & \unsup & ein aufgeregter turner in einem blau-rosa \textbf{outfit} führt eine band aus . \\
\hline
\end{tabular}}}
\end{center}
\caption{\label{tab:amb-samples} Samples of ambiguous words translation on \B{2016} for both EN-FR and EN-DE. In both cases more correct translations are provided by \unsup. Bold highlights target words and their translations.
}
\end{table*}
 
\begin{table*}[t!]
\begin{center}
\scalebox{1.0}{
\small{
\begin{tabular}{c||l|p{11.5cm}}
\hline
Freq. 1 & source word & traveler \\
   & gold target word & reisender\\
   & source sentence & an oriental \textbf{traveler} awaits his turn at the currency exchange  .\\
   & reference sentence & ein orientalischer \textbf{reisender} `traveler' wartet am wechselschalter bis er dran ist .\\
\cline{2-3}
   & MLE & ein orientalisch aussehender \textbf{behinderter} `disabled' wartet darauf , dass die kurve sich die glastür aufhebt .\\
   & \super & ein orientalisch aussehender \textbf{techniker} `technician' wartet auf die hecke seiner kurve .\\
   & \unsup & ein orientalisch aussehender \textbf{mann} `man' wartet darauf , dass seine kurve auf den fehenk die kurve ist .\\
\hline
Freq. 28 & source word & check \\
   & gold target word & scheck \\
   & source sentence & a woman is holding a large \textbf{check} for kids food basket .\\
   & reference sentence & eine frau hält einen großen \textbf{scheck} `check' für " kids' food basket " .\\
\cline{2-3}
   & MLE & eine frau hält ein großes \textbf{überprüfen} `proof' für kinder .\\
   & \super & eine frau hält einen großen \textbf{informationen} `information' für kinder in den korb .\\
   & \unsup & eine frau hält ein großes \textbf{überprüfen} `proof' für kinder , die einen korb zu verkaufen ist .\\
\hline
\end{tabular}}}
\end{center}
\caption{\label{tab:freq-samples} Samples of translations for words of different frequency on \B{2016} EN-DE. In both cases more correct translations are provided by \unsup. Bold highlights target words and their translations.
}
\end{table*}
 
\subsection{Qualitative analysis}\label{ssec:results-human-analysis}

To get further insights into the general results, we also performed human evaluation of the outputs for MLE, \super,  and \unsup using professional in-house expertise. This was done for \B{COCO} EN-FR and \B{2016} EN-DE as two sets with contrastive results in the lexical translation experiment. 
 
For this human analysis, we randomly selected test samples (50 samples per language pair per group) with source words of different frequency in the training data: rare words (frequency 1) and other words (frequency $\geq$ 10). These other words are randomly chosen from the sentences that differ in their translation across setups. The resulting average frequency of those words is around 40 for both language pairs. A rank of quality (both fluency and adequacy together) is assigned by the human evaluator from 1 to 3, allowing ties. Following the common practice in MT, each system was then assigned a score which reflects how often it was judged to be better or equal to other systems~\cite{bojar-etal-2017-findings}.

\begin{table}[!h]
\begin{center}
\scalebox{0.72}{
\begin{tabular}{c c c c c}
\toprule
Lang & Words & MLE & \super & \unsup \\ \midrule
\multirow{2}{*}{EN-FR}
& Rare (Freq. 1) & 1.76 & \bf 1.88 & 1.68 \\
& Other & \bf 1.88 & 1.82 & 1.86 \\
\midrule
\multirow{2}{*}{EN-DE}
& Rare (Freq. 1) & 1.72 & \bf 1.74 & 1.70 \\
& Other & 1.93 & 1.83 & \bf 1.94 \\
\bottomrule
\end{tabular}}
\end{center}
\caption{\label{table:human_res} Human ranking results for \B{2016} EN-DE and \B{COCO} EN-FR test set. Bold highlights best results per group of word types. The first column indicates the groups of word types. Results are averaged for all words per word type group.}
\end{table}

Results are in Table~\ref{table:human_res}. We observe a tendency of \super to do well on the translation of rare source words, but not so well on the translation of words in the middle frequency range (this observation is confirmed by the analysis of the frequency of output words, see~Appendix~\ref{subsec:distrib_analysis}, see Table~\ref{tab:freq-samples}). Our unsupervised reward tends to increase the performance on more frequent words (`Other' in~Table~\ref{table:human_res}) by promoting their less common translations in the distribution, hence better translations for ambiguous words from our previous experiment. These ambiguous words are quite frequent, they potentially have multiple possible translations but only one correct translation in a given context. 

\section{Conclusions} \label{sec:concl}

We propose and reformulate SAC reinforcement learning approaches to help machine translation through better exploration and less reliance on the reward function. To provide a good trade-off between exploration and quality, we devise two reward methods in the supervised and dynamic unsupervised manner. The maximum entropy off-policy SAC algorithm  mitigates the overfitting problem when evaluated in the out-of-domain space; both rewards introduced in our SAC architecture can achieve better quality for lexical translation of ambiguous words, particularly the rare senses of words. The formulation of the unsupervised reward and its potential to influence translation quality open perspectives for future studies on the subject. We leave the exploration of how those supervised and unsupervised rewards could be combined to improve MT for future work.

\section*{Acknowledgments}
The authors thank the anonymous reviewers for their useful feedback. This work was supported by the MultiMT (H2020 ERC Starting Grant No. 678017) and the Air Force Office of Scientific Research (under award
number FA8655-20-1-7006) projects. Marina Fomicheva and Lucia Specia were supported by funding from the Bergamot project (EU H2020 grant no. 825303). We also thank the annotators for their valuable help.

\bibliography{anthology,aacl-ijcnlp2020}
\bibliographystyle{acl_natbib}

\clearpage
\appendix

\section{Training Details}
\label{appsec:training}

\subsection{Hyperparameters}
\label{appen:hyper}

For the NMT RNN agent, the dimensions of embeddings and GRU hidden states are set to 200 and 320, respectively. The decoder's input and output embeddings are shared~\cite{press-wolf-2017-using}. We use Adam~\cite{kingma2014adam} as the optimiser and set the learning rate and mini-batch size to 0.0004 and 64, respectively. A weight decay of $1e\rm{-}5$ is applied for regularisation. We clip the gradients if the norm of the full parameter vector exceeds $1$~\cite{pascanu2013difficulty}. The four Q-networks are identical to the agent. 

For the unsupervised reward setting, we use 2 two-layer feed-forward neural network (both dimensionalities are equal to 100). We use again Adam as the optimiser and set the learning rate and mini-batch size to 0.0001 and 64, respectively.

\begin{table}[!h]
	\centering
	\begin{tabular}{l|c}
		\toprule
		\bf Hyper-parameters & \\
		\midrule
		{\bf Pre-train Critic}  &\\
		\midrule
		optimiser & Adam \\
		learning rate & 0.0003 \\
		batch size & 64 \\
		$\tau$ (target net speed)  & 0.005 \\
		$\alpha$ (entropy regularization)  & 0.001 \\
		buffer size  & 1000 \\
		length penalty & 0.0001 \\
		\midrule
		\bf Joint Training \\
		\midrule
		optimiser & Adam \\
		learning rate & 0.0004 \\
		batch size & 64 \\
		$\tau$ (target net speed)  & 0.005 \\
		$\alpha$ (entropy regularization)  & 0.001 \\
		buffer size  & 1000 \\
		length penalty & 0.0001 \\
	    $\lambda_{MLE}$  & 0.1 \\
		\bottomrule
	\end{tabular}
	\caption{Hyper-parameters for SAC training.}
	\label{tab:hyperparam-erac}
\end{table}

\subsection{Training} 
We use PyTorch~\cite{paszke2017automatic} (v1.4, CUDA 10.1) for our experiments. We early stop the actor training if validation loss does not improve for 10 epochs, we pretrain critics for 5 epochs for the \B{Multi30K} datasets and for 3 epochs for the larger \B{IWSLT 2014}. We early stop the SAC training if validation BLEU does not improve for 10 epochs. For all the setups, we also halve the learning rate if no improvement is obtained for two epochs. On a single NVIDIA RTX2080-Ti GPU, it takes around 5-6 hours up to 36 hours to train a model depending on the data size and the language pair. The number of learnable parameters is about 7.89M for smaller Multi30K models and about for 15.64M for the bigger IWSLT model. All models were re-trained 3 times to ensure reproducibility.
 
\subsection{Soft Actor-Critic Training Algorithm} 
We describe the main steps of SAC training in Algorithm~\ref{alg:sac}.\\

\begin{algorithm}
\SetAlgoLined
  Initialise parameters: \\
  Q function: $\theta$; \\
  Policy: $\phi$;\\
  Unsupervised Reward: $\delta$;\\
  Replay Buffer: $\mathcal{D}$ $\leftarrow$ $\emptyset$;\\
 \For{each iteration}{
  \For{each translation step}{
    $a_{t}$ $\sim$ $\pi_{\phi}(a_t, s_t)$;\\
    $s_{t+1}$ $\sim$ $p(s_{t+1} | s_t, a_t)$;\\
    $\mathcal{D}$ $\leftarrow$ $\mathcal{D}$ $\cup$ $\{s_t, a_t, r(s_t, a_t)$, $s_{t+1}\}$ ;\\ 
   }
  \For{each gradient step}{
    $\theta_i \leftarrow \theta_i - \lambda_{Q}$ $\nabla_{\theta_i}L(\theta_i)\ \text{for}\ i \in \{1, 2\}$;\\
    $\phi \leftarrow \phi - \lambda_{\pi} \nabla_{\phi}J(\phi)$;\\
    $\alpha \leftarrow \alpha - \lambda_{\pi}$ $\nabla_{\alpha}J(\alpha)$;\\
    $\theta_i \leftarrow \tau\theta_{i} + (1-\tau)\bar{\theta_i}$\\ $\text{for}\ i \in \{1, 2\}$;\\
    \If{unsupervised reward}{
       $\delta \leftarrow \delta - \lambda_{z} \nabla_{\delta}r(\delta)$;\\ 
    }
  }
 }
 \caption{Soft Actor-Critic.}
 \label{alg:sac}
\end{algorithm}

\subsection{Domain Distance}\label{appssec:domain}

To assess to what extent the test sets used in our experiments can be considered out-of-domain, we train (i) an English language model on \B{Multi30K} training set; and (ii) a German language model on the \B{IWSLT 2014} training set.\footnote{We train Transformer language models using the fairseq toolkit~\cite{ott2019fairseq}.} Table \ref{tab:domain_distance} shows language model perplexities on the Mutli30k test data. With respect to the \B{IWSLT 2014} model, \B{Multi30K} test sets are clearly very different from the training data. 
With respect to the Multi30K model, \B{2017} and \B{COCO} are more distant from the train partition than 2016 testset. 

\begin{table}[!h]
\begin{center}
\scalebox{1.0}{
\begin{tabular}{l c c c}
\toprule
LM & \B{2016} & \B{2017} & \B{COCO} \\
\midrule
\B{Multi30K} & 44.07 & 79.95 & 77.7 \\
\B{IWSLT 2014} & 579.47 & 403.54 & 381.56\\
\bottomrule
\end{tabular}}
\end{center}
\caption{Perplexity on \B{Multi30K} testsets for \B{Multi30K} and \B{IWSLT 2014} language models.}
\label{tab:domain_distance}
\end{table}

\begin{table*}[!ht]
\begin{center}
\scalebox{0.85}{
\begin{tabular}{lc@{\hspace{1cm}}lll@{\hspace{0.5cm}}lll@{\hspace{0.5cm}}lll@{\hspace{0.5cm}}}
\toprule
& & \multicolumn{3}{c}{\B{2016}} & \multicolumn{3}{c}{\B{2017}} & \multicolumn{3}{c}{\B{COCO}} \\ 
\cline{3-11}
& model & BLEU & METEOR & TER & BLEU & METEOR & TER & BLEU & METEOR & TER \\ \midrule
\multirow{2}{*}{\rotatebox{90}{EN-FR}}
& MLE & 58.8 & 73.8 & \bf 26.7 & 54.2 & 70.2 & 30.1 & 42.6 &  62.1 & 36.0  \\
& \super & \bf 59.4 & \bf 74.0 & \bf 26.7* & \bf 55.2 & \bf 70.8 & \bf 29.2 & \bf 44.1 & \bf 63.4 & \bf 35.5 \\
& \unsup &  58.2 & 73.6 & 27.3 & 54.4* & 70.6 & 29.8 & 43.5 & 63.2  & 35.7* \\
\midrule
\multirow{2}{*}{\rotatebox{90}{EN-DE}}
& MLE & \bf 37.5 & 56.3  & \bf 42.1  & \bf 33.8 & \bf 53.1  & \bf 47.6  & 29.3 &  \bf 49.3 & \bf 50.9\\
& \super & 36.6 & 56.2* & 43.2 & 33.5 & \bf 53.1* & \bf 47.6* & \bf 29.6* & \bf 49.3* & 51.0* \\
& \unsup & 36.3 & \bf 56.5*  & 44.1  & 33.1  & 52.9* & 48.7  & 28.3  & 48.6 & 51.5\\
\bottomrule
\end{tabular}
}
\end{center}
\caption{\label{table:amb-bleu} Results on the test sets for ambiguous words.}
\end{table*}

\subsection{Analysis of distributions}\label{subsec:distrib_analysis}

\begin{figure}
\includegraphics[width=0.5\textwidth]{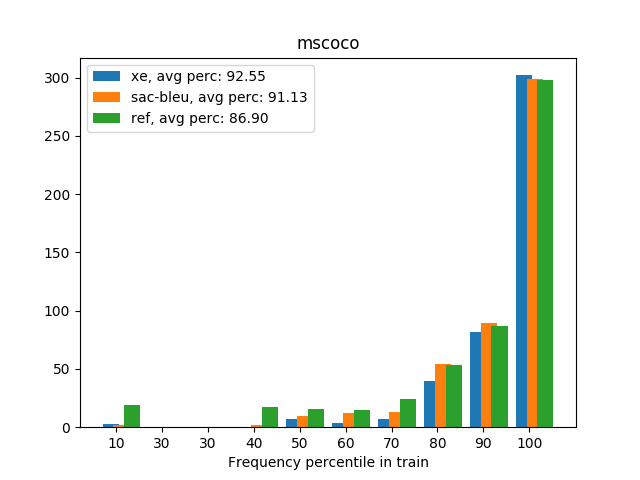}
\caption{Training frequency for \B{COCO} words as translated by MLE and \super. We also report reference frequencies.}
\label{fig:frequencies}
\end{figure}

We argue that the improvement over MLE can be partially attributed to a better handling of less frequent words. It has been shown that rare words tend to be under-represented in NMT \cite{koehn-knowles-2017-six, shen-etal-2016-minimum}. RL training with regularized entropy might mitigate this issue due to a better exploration of the action space. To illustrate this point, we compute the training frequency of the words generated by the NMT systems for the sentences where an improvement over MLE is observed. Figure \ref{fig:frequencies} shows the training frequency percentiles for MLE and \super English-French translations of the \B{COCO} testset. Reference frequencies are also provided for comparison. We observe that although both MLE and SAC contain more frequent words than the reference, this tendency is less pronounced for SAC. \red{We relate this observation to the fact that our SAC outperforms MLE for the ambiguous word translation (Table~\ref{tab:lex_translation}) where the most frequent translation is not always the correct one.}

\end{document}